\documentclass[sn-chicago, iicol]{sn-jnl}

\usepackage{graphicx}%
\usepackage{multirow}%
\usepackage{amsmath,amssymb,amsfonts}%
\usepackage{amsthm}%
\usepackage{mathrsfs}%
\usepackage[title]{appendix}%
\usepackage{xcolor}%
\usepackage{textcomp}%
\usepackage{manyfoot}%
\usepackage{booktabs}%
\usepackage{algorithm}%
\usepackage{algorithmicx}%
\usepackage{algpseudocode}%
\usepackage{listings}%

\usepackage{amsmath}
\usepackage{amssymb}
\newcommand{\argmax}{\mathop{\rm arg~max}\limits}
\usepackage{subcaption}
\usepackage{booktabs}
\usepackage{colortbl}
\usepackage{xcolor} 
\usepackage{bbding}
\usepackage{pifont}
\usepackage{arydshln}

\newcommand{\eg}{\textit{e.g., }}
\newcommand{\ie}{\textit{i.e., }}
\newcommand{\etal}{\textit{et al. }}
\usepackage[whole]{bxcjkjatype}

\theoremstyle{thmstyletwo}%

\theoremstyle{thmstylethree}%

\begin{document}

\title[Article Title]{GL-MCM: Global and Local Maximum Concept Matching \\for Zero-Shot Out-of-Distribution Detection}


\author*[1]{\fnm{Atsuyuki} \sur{Miyai}}\email{miyai@hal.t.u-tokyo.ac.jp}

\author[1,2]{\fnm{Qing} \sur{Yu}}\email{yu@hal.t.u-tokyo.ac.jp}

\author[3]{\fnm{Go} \sur{Irie}}\email{goirie@ieee.org}

\author[1]{\fnm{Kiyoharu} \sur{Aizawa}}\email{aizawa@hal.t.u-tokyo.ac.jp}

\affil[1]{\orgname{The University of Tokyo}}
\affil[2]{\orgname{LY corporation}}
\affil[3]{\orgname{Tokyo University of Science}}

\abstract{

Zero-shot out-of-distribution (OOD) detection is a task that detects OOD images during inference with only in-distribution (ID) class names.
Existing methods assume ID images contain a single, centered object, and do not consider the more realistic multi-object scenarios, where both ID and OOD objects are present. To meet the needs of many users, the detection method must have the flexibility to adapt the type of ID images. 
To this end, we present Global-Local Maximum Concept Matching (GL-MCM), which incorporates local image scores as an auxiliary score to enhance the separability of global and local visual features. Due to the simple ensemble score function design, GL-MCM can control the type of ID images with a single weight parameter. Experiments on ImageNet and multi-object benchmarks demonstrate that GL-MCM outperforms baseline zero-shot methods and is comparable to fully supervised methods. Furthermore, GL-MCM offers strong flexibility in adjusting the target type of ID images. The code is available via \url{https://github.com/AtsuMiyai/GL-MCM}.
}




\maketitle
\section{Introduction}\label{sec:intro}

Detecting out-of-distribution (OOD) samples is essential for deploying machine learning models in real-world environments, where new classes of data can naturally emerge and necessitate caution. Common OOD detection approaches utilize single-modal supervised learning~\citep{hendrycks2016baseline, hendrycks2019scaling, huang2021importance, liang2017enhancing, sun2022out, liu2020energy, lee2018simple, haoqi2022vim, arora2021types}. These supervised methods achieve good results but have limitations \textit{e.g.}, these methods require a lot of computational and annotation costs for training. These days, CLIP~\citep{radford2021learning} achieves surprising performances for various downstream tasks, and its application to OOD detection is drawing increasing interest~\citep{esmaeilpour2022zero,ming2022delving, tao2023non}

\begin{figure}[t]
  \centering
  \includegraphics[keepaspectratio, scale=0.22]
  {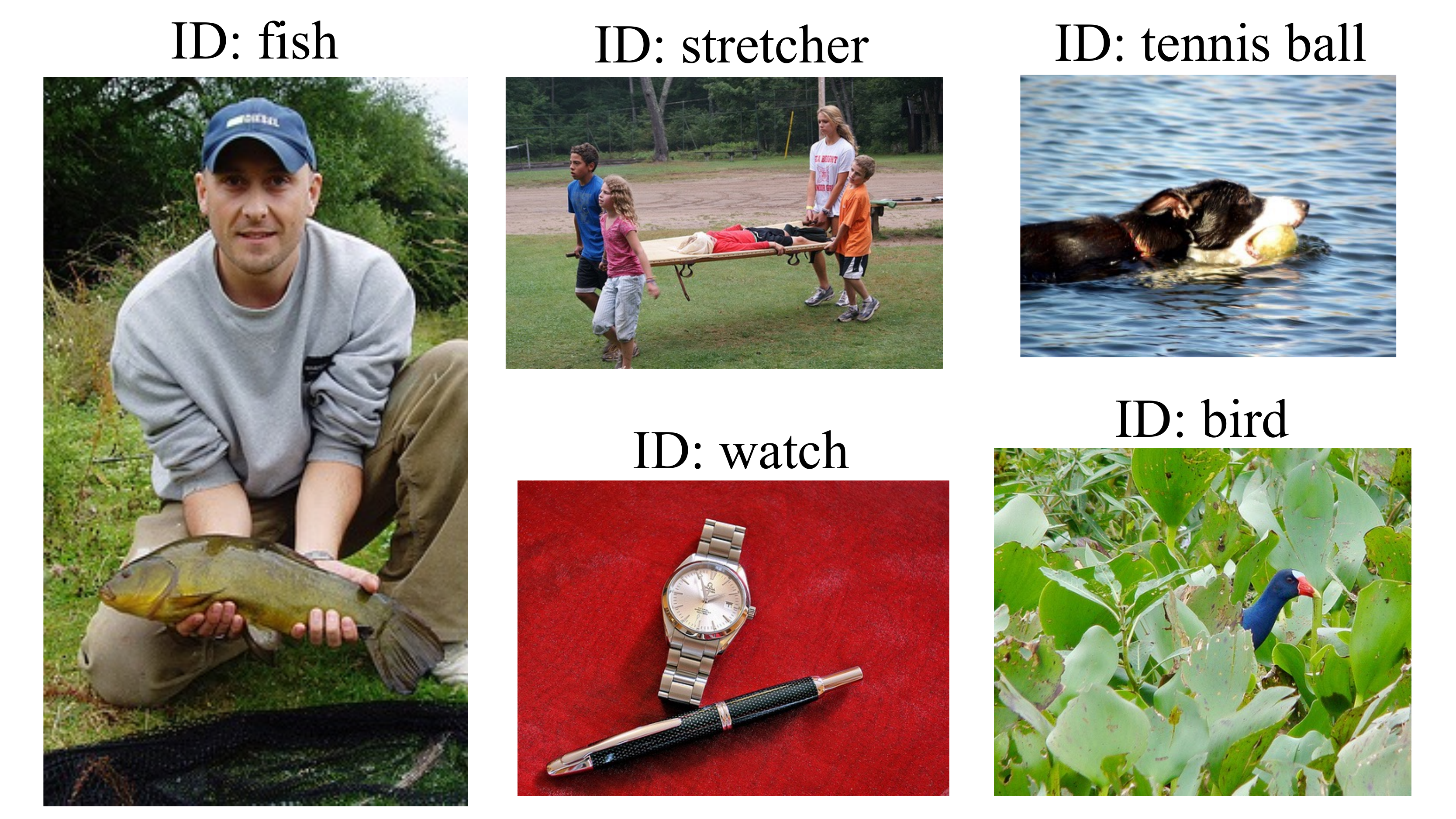}
  \caption{\small{\textbf{Multi-object images in ImageNet.} ImageNet contains numerous images with multiple objects. It reflects the reality that many real-world photos do not feature a single ID object, and, instead, they often include OOD objects within the frame.
  }}
  \label{fig:imagenet_multi_object}
\end{figure}

While some CLIP-based OOD detection methods have been proposed~\citep{fort2021exploring, esmaeilpour2022zero, ming2022delving, jiang2024negative, cao2024envisioning}, existing approaches overlook the characteristics of ID images. Real-world images are rarely clean, centered compositions of a single object. Instead, they frequently contain multiple objects, with non-ID objects occasionally present within the scene. An example of an ImageNet image with multiple objects is shown in Fig.~\ref{fig:imagenet_multi_object}. As illustrated, real-world images commonly include several objects. 

Whether images featuring ID and OOD objects are detected as ID or not depends on downstream tasks and the user's needs. For example, most images in the ``Stretcher" category include people alongside the stretcher, so treating only single-object images as ID is impractical for real-world applications.
Therefore, it is important to propose a method to detect images as ID even if they have both ID and OOD objects. Furthermore, ideally, it is crucial for the detection method to have the flexibility to change the type of ID images in a way that meets their specific needs.

In this paper, we propose Global-Local Maximum Concept Matching (GL-MCM), a simple, effective, and flexible approach that leverages local information from images as an auxiliary score. CLIP's global image features, commonly used in existing method~\citep{ming2022delving}, have the mixed concept of ID and OOD objects and ID object features are contaminated by OOD object features. On the other hand, CLIP's local feature would have information on all objects in the image. We first introduce Local Maximum Concept Matching (L-MCM), which applies the softmax scaling to the local visual-text alignments to enhance the separability of the local features. With L-MCM, we propose Global-Local Maximum Concept Matching (GL-MCM), which incorporates both global and local softmax-scaled scores. GL-MCM can compensate for the shortcomings of both global and local visual-text alignments, which can detect any image containing ID objects from OOD images regardless of whether ID objects appear globally or locally.

Because of its simplicity, GL-MCM offers a variety of user-friendly advantages as well as performance improvements. Firstly, users can control the type of ID images (ID-centered image or multi-object image) that they aim to detect by changing just a single weighting parameter of the local and global scores. Secondly, due to its high scalability, GL-MCM boosts the performance of few-shot learning methods~\citep{miyai2023locoop, zhang2022tip}.

We evaluated our methods on common ImageNet OOD benchmarks~\citep{huang2021mos} and our curated multi-object benchmark (MS-COCO~\citep{lin2014microsoft} and Pascal-VOC~\citep{everingham2009pascal}). Our experimental results indicate that GL-MCM outperforms comparison zero-shot methods and is comparable to fully supervised methods (Table~\ref{table:single_IN} and \ref{table:coco_voc}). Also, GL-MCM boosts few-shot learning methods (Table~\ref{table:single_IN}). 
Additionally, due to its simplicity, GL-MCM provides high flexibility for adjusting the target type of ID images (Fig.~\ref{fig:ablation_lambda}) and Fig.~\ref{fig:histogram}).

The contributions of our paper are summarized as follows:
\begin{itemize}
      \item \textbf{Proposal of Global-Local Maximum Concept Matching (GL-MCM)}: We propose GL-MCM, a simple and flexible method based on global and local vision-language concept alignments. GL-MCM can detect ID images even if ID objects appear globally or locally.
      \item \textbf{Improved Flexibility for ID Image Detection}: GL-MCM allows users to adjust the balance between global and local scores, providing flexibility to change the types of ID images (ID-dominant or multi-object ID images) according to specific application needs.
      \item \textbf{Enhanced Performance in Various Benchmarks}: GL-MCM outperforms existing methods in common ImageNet benchmarks and multi-object benchmarks. Also, it boosts the performance of few-shot methods. 
\end{itemize}

\begin{figure*}[t]
  \centering
  \includegraphics[keepaspectratio, scale=0.45]
  {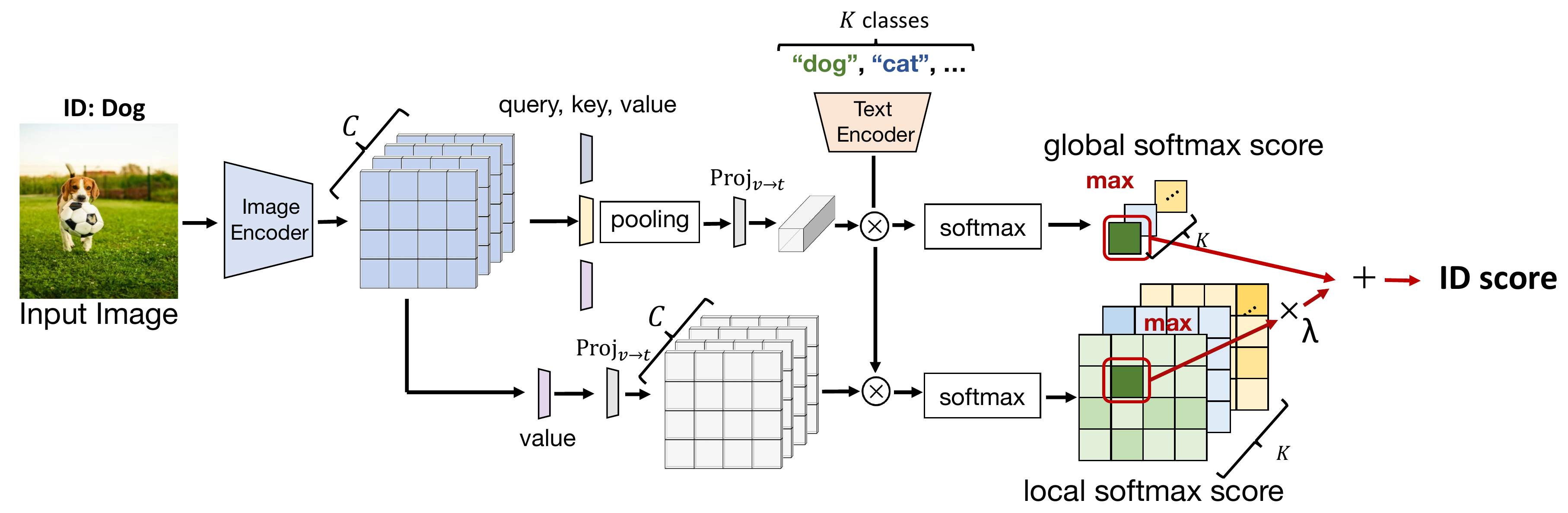}
  \caption{\textbf{Overview of the Global-Local Maximum Concept Matching (GL-MCM) framework.} Our approach utilizes both global and local softmax scores to calculate the ID confidence. By incorporating both global and local scores, our framework compensates for the respective weaknesses of the global and local alignments.}
  \label{fig:overview_GL-MCM}
\end{figure*}

\section{Methodology}
\subsection{Problem Statements}
\label{subsec:problem_statement}
Zero-shot OOD detection is a task to detect OOD images with only ID class names, not ID images~\citep{fort2021exploring, esmaeilpour2022zero, ming2022delving}.
In zero-shot OOD detection, the ID classes refer to the classes we aim to classify, which are different from the classes of the upstream pre-training. Accordingly, OOD classes are the classes that do not belong to any of the ID classes. OOD detector can be viewed as a binary classifier that identifies whether the image is an ID image or an OOD image. 
\subsection{Review of MCM}
\label{preliminaries_MCM}
\textbf{Overview of MCM.}
MCM~\citep{ming2022delving} is a state-of-the-art zero-shot OOD detection method. It calculates the confidence scores with the softmax score of the similarity between global image features and class textual features. For CLIP's image encoder (\textit{i.e.}, modified ResNet), the original attention pooling layer pools the visual feature map first and then projects the global feature vector into text space by $\operatorname{Proj}_{v \rightarrow t}$. 
We define the visual and textual encoders as $E_v(\cdot)$ and $E_t(\cdot)$. For any test input image, we define the output feature map of $E_v(\cdot)$ as $\mathbf{x}\in \mathbb{R}^{HW\times C}$, where $H$, $W$, and $C$ are the height, width and the number of channels of the feature maps. The detailed operation for creating the global feature $\mathbf{x^{\prime}} \in \mathbb{R}^{C}$ is as follows:
\small
\begin{equation}
\begin{aligned}
\mathbf{x^{\prime}} &= \operatorname{AttnPool}(\mathbf{x})
\\ &=
\operatorname{Proj}_{v \rightarrow t}\left(\sum_i \operatorname{softmax}\left(\frac{q(\bar{\mathbf{x}}) k\left(\mathbf{x}_i\right)^T}{D}\right) \cdot v\left(\mathbf{x}_i\right)\right),
\label{mcm_global_feature}
\end{aligned}
\end{equation}
\normalsize
where $q$, $v$, and $k$ denote the query, value, and key, and they are independent linear embedding layers in CLIP. $D$ is a constant scaling factor. $\bar{\mathbf{x}}\in \mathbb{R}^{C}$ is created by applying global average pooling to $\mathbf{x}$. $\mathbf{x}_i \in \mathbb{R}^{C}$ denotes the visual feature of each region $i$ of feature map $\mathbf{x}$. 

Let $\mathcal{T}_{\text{in}}$ denote the set of test prompts containing $K$ class labels (\textit{e.g.},``a photo of a [CLASS]''). The output text feature for $t \in \mathcal{T}_{\text{in}}$ is defined as  $\mathbf{y}_t = E_t(t)$.
The score function of MCM is defined as follows:
\begin{equation}
S_{\mathrm{MCM}} =\max _{t  \in \mathcal{T}_{\text{in}}} \frac{e^{\operatorname{sim}\left(\mathbf{x^{\prime}}, \mathbf{y}_t\right) / \tau}}{\sum_{c\in  \mathcal{T}_{\text{in}}} e^{\operatorname{sim}\left(\mathbf{x^{\prime}}, \mathbf{y}_c\right) / \tau}},
\end{equation}
where $\operatorname{sim}(\mathbf{u_1}, \mathbf{u_2}) = \mathbf{u_1}\cdot\mathbf{u_2}/\|\mathbf{u_1}\|\cdot\|\mathbf{u_2}\|$ denote cosine similarity between $\mathbf{u_1}$ and $\mathbf{u_2}$ and $\tau$ is the temperature. \cite{ming2022delving} stated that softmax scaling improves the separability between ID and OOD images.
\\[2mm]
\noindent\textbf{Limitations of MCM.}
In Eq.~(\ref{mcm_global_feature}), $\sum_i \operatorname{softmax}\left(\frac{q(\bar{\mathbf{x}}) k\left(\mathbf{x}_i\right)^T}{D}\right)$ is the attention map, which aggregates the feature map into the global feature. The major problem here is that the attention map in the pooling operation has no knowledge of which objects in the image are ID objects or OOD objects and has overall knowledge of the whole image. 
Therefore, when the OOD object appears in the image, its global feature has less ID information, and MCM incorrectly judges the image as an OOD image.

\subsection{Proposed Approach}
\label{proposed_approach}
In the proposed approach, we leverage the local embeddings of CLIP. In the original attention pooling layer in CLIP, we can replace the projection into text space $\operatorname{Proj}_{v \rightarrow t}$ and the pooling operation as follows:
\scriptsize	
\begin{equation}
\begin{aligned}
&\operatorname{AttnPool}(\mathbf{x})\\&=\operatorname{Proj}_{v \rightarrow t}\left(\sum_i \operatorname{softmax}\left(\frac{q(\bar{\mathbf{x}}) k\left(\mathbf{x}_i\right)^T}{D}\right) \cdot v\left(\mathbf{x}_i\right)\right)\\
&=\sum_i \operatorname{softmax}\left(\frac{q(\bar{\mathbf{x}}) k\left(\mathbf{x}_i\right)^T}{D}\right) \cdot \operatorname{Proj}_{v \rightarrow t}(v\left(\mathbf{x}_i\right)) \\
&= \operatorname{Pool}(\operatorname{Proj}_{v \rightarrow t}(v\left(\mathbf{x}_i\right))),
\label{mcm_global_feature_modified}
\end{aligned}
\end{equation}
\normalsize
where $q$, $v$, and $k$ are independent linear embedding layers and $\operatorname{Pool}(\cdot)$ denotes $\sum_i \operatorname{softmax}\left(\frac{q(\bar{\mathbf{x}}) k\left(\mathbf{x}_i\right)^T}{D}\right)$.

By removing the pooling operation in Eq.~(\ref{mcm_global_feature_modified}), we can project the visual feature $\mathbf{x}_i$ of each region $i$ to the textual space~\citep{zhou2022extract}:
\begin{equation}
\begin{aligned}
\mathbf{x^{\prime}}_i = \operatorname{Proj}_{v \rightarrow t}(v\left(\mathbf{x}_i\right)).
\label{local_clip}
\end{aligned}
\end{equation}
For ViT, we can also obtain this feature map with similar procedures~\citep{zhou2022extract}. This feature has a rich local visual and textual alignment~\citep{zhou2022extract}. However, this local feature map calculates the similarity on a region-by-region basis, so it might be difficult to calculate the accurate similarity to the global object appearing over many regions. Therefore, we propose a simple yet effective method called Global-Local Maximum Concept Matching (GL-MCM), which incorporates both global and local features. In Fig.~\ref{fig:overview_GL-MCM}, we show the overview of GL-MCM. For local alignments, we extend the concepts of MCM for local alignment features, and we first propose Local Maximum Concept Matching (L-MCM), in which we characterize ID confidences by the closest distance between the local visual text concept similarities.
The score function for local matching is defined as follows:
\begin{equation}
S_{\mathrm{L-MCM}} =\max _{t, i} \frac{e^{\operatorname{sim}\left(\mathbf{x^{\prime}}_i, \mathbf{y}_t\right) / \tau}}{\sum_{c\in  \mathcal{T}_{\text{in}}} e^{\operatorname{sim}\left(\mathbf{x^{\prime}}_i, \mathbf{y}_c\right) / \tau}}.
\label{eq:l_mcm}
\end{equation}
Note that $\tau$ is the temperature and, unlike common segmentation tasks~\citep{zhou2022extract}, the softmax scaling with $\tau$ is crucial to improve the separability between ID and OOD. This is inspired by the theoretical analysis that the softmax scaling for global features enhances the separability between ID and OOD images~\citep{ming2022delving}.

Then, we propose GL-MCM, which ensembles the score functions of MCM for global objects and L-MCM for local objects.
The score function for GL-MCM is defined as follows:
\begin{equation}
S_{\mathrm{GL-MCM}} = S_{\mathrm{MCM}} +  \lambda S_{\mathrm{L-MCM}}.
\label{eq:gl_mcm}
\end{equation}
Here, $\lambda$ is a hyperparameter that allows users to control what kind of images they aim to detect as ID. 

The small value of $\lambda$ enables users to detect the images with dominant ID objects and the high value enables users to detect images with both the ID and OOD objects. Note that the principle of our method is not limited to the use of CLIP's pure local features~\citep{zhou2022extract}, and can generally be applicable for other latest localization methods~\citep{Xu_2023_CVPR}.

\noindent\textbf{Remarks on the Novelty of GL-MCM.} Readers familiar with OOD detection might consider that this is a naive extension of MCM. However, we argue that the ability to separate ID and OOD with CLIP's local features has not been explored at all. L-MCM bridges the fields of CLIP-based localization and OOD detection, integrating these two previously disparate fields. Hence, the novelty of GL-MCM is not measured by the complexity, and we consider GL-MCM to be a solid technical novelty. 

\begin{table*}[t]
\centering
\footnotesize
\centering
     {\tabcolsep = 0.5mm
    \begin{tabular}{@{}lcccccccccccccc@{}} \toprule
    & \multicolumn{2}{c}{iNaturalist} && \multicolumn{2}{c}{SUN} && \multicolumn{2}{c}{Places}  && \multicolumn{2}{c}{Texture} && \multicolumn{2}{c}{\textbf{Average}}\\ 
    \cmidrule(lr){2-3} \cmidrule(lr){5-6} \cmidrule(lr){8-9} \cmidrule(lr){11-12} \cmidrule(lr){14-15}
   \textbf{Method} & FPR95$\downarrow$ & AUROC$\uparrow$ && FPR95$\downarrow$ & AUROC$\uparrow$ && FPR95$\downarrow$ & AUROC$\uparrow$ && FPR95$\downarrow$ & AUROC$\uparrow$ && FPR95$\downarrow$ & AUROC$\uparrow$ \\ \midrule
    &&\multicolumn{11}{c}{\hspace{20pt} \textbf{Fully-supervised} (reference)}\\
    Fort~\etal/MSP (ViT-B)$^\dagger$ & 15.07 & 96.64 && 54.12 & 86.37 &&  57.99 & 85.24 && 53.32 & 84.77 &&  45.12 & 88.25 \\
    Energy$^\dagger$ \; \; \; & 21.59 & 95.99 &&  34.28 & 93.15 &&  36.64 & 91.82 && 51.18 & 88.09 && 35.92 & 92.26 \\
    ODIN$^*$ & 30.22 & 94.65 && 54.04 & 87.17 && 55.06 & 85.54 && 51.67 & 87.85 && 47.75 & 88.80 \\
    ViM$^*$ & 32.19 & 93.16 && 54.01 & 87.19 && 60.67 & 83.75 && 53.94 & 87.18 && 50.20 & 87.82\\
    KNN~\citep{sun2022out}$^*$ & 29.17 & 94.52 && 35.62 & 92.67 && 39.61 & 91.02 && 64.35 & 85.67 && 42.19 & 90.97 \\
    NPOS~\citep{tao2023non}$^*$ & 16.58 & 96.19 && 43.77 & 90.44 && 45.27 & 89.44 && 46.12 & 88.80 && 37.93 & 91.22 \\
    &
    &\multicolumn{10}{c}{\hspace{0pt} \textbf{Zero-shot}}\\
    MCM~\citep{ming2022delving} & 30.91 & 94.61 && 37.59 & 92.57 && 44.69 & 89.77 && 57.77 & 86.11 && 42.74 & 90.77 \\
    \rowcolor[gray]{0.90}
    L-MCM (ours) & 49.19 & 86.96 && 73.65 & 76.62 && 79.14 & 71.86 && 92.39 & 59.48 && 73.59 & 73.73 \\     
    \rowcolor[gray]{0.90}
     GL-MCM (ours) & 17.42 & 96.44 && 30.75 & 93.44 && 37.62 & 90.63 && 55.20 & 85.54 && 35.25 & 91.51 \\  
     && \multicolumn{11}{c}{\hspace{0pt} \textbf{Few-shot} (16-shot)}\\
    CoOp\textsubscript{MCM} & 28.00 & 94.43 && 36.95 & 92.29 && 43.03 & 89.74 && 39.33 & 91.24 && 36.83 & 91.93 \\
    \rowcolor[gray]{0.90}
    CoOp\textsubscript{GL} (ours) & 16.87 & 96.28 && 28.27 & 93.19 && 36.07 & 90.61 && \textbf{38.92} & 90.29 && 30.03 & 92.59 \\
    LoCoOp\textsubscript{MCM} & 23.06 & 95.45 && 32.70 & 93.35 && 39.92 & 90.64 && 40.23 & \textbf{91.32} && 33.98 & 92.69 \\
     \rowcolor[gray]{0.90}
     LoCoOp\textsubscript{GL} (ours) & \textbf{16.79} & \textbf{96.60} && \textbf{24.66} & \textbf{94.76} &&	 \textbf{33.64} & \textbf{91.87} && 39.63 & 91.14 && \textbf{28.68} & \textbf{93.59} \\
    \bottomrule
    \end{tabular}
    }
\caption{\small{\textbf{Main results on ImageNet benchmarks.} Methods $\dagger$ are fine-tuned on ImageNet-1K and cited from~\citep{ming2022delving}. Methods $^*$ are fine-tuned on ImageNet-1K and cited from~\citep{tao2023non}. GL-MCM outperforms MCM and is comparable to supervised methods on ImageNet. Also, GL-MCM boosts the performance of few-shot learning methods.}}
\label{table:single_IN}
\end{table*}

\begin{table*}[t]
\centering
\footnotesize
\centering
\begin{minipage}[t]{1.0\linewidth}
\centering
     {\tabcolsep = 0.2mm
    \begin{tabular}{@{}lccccccccccccccccc@{}} \toprule
    & \multicolumn{2}{c}{iNaturalist} && \multicolumn{2}{c}{SUN}  && \multicolumn{2}{c}{Texture} && \multicolumn{2}{c}{IN-22K} && \multicolumn{2}{c}{VOC} && \multicolumn{2}{c}{\textbf{Average}}\\ 
    \cmidrule(lr){2-3} \cmidrule(lr){5-6} \cmidrule(lr){8-9} \cmidrule(lr){11-12} \cmidrule(lr){14-15} \cmidrule(lr){17-18}
   \textbf{Method} & FPR95$\downarrow$ & AUROC$\uparrow$ && FPR95$\downarrow$ & AUROC$\uparrow$ && FPR95$\downarrow$ & AUROC$\uparrow$ && FPR95$\downarrow$ & AUROC$\uparrow$ && FPR95$\downarrow$ & AUROC$\uparrow$ && FPR95$\downarrow$ & AUROC$\uparrow$ \\ \midrule
     MCM & 64.42 & 90.75 &&  84.18 & 81.81 &&  \textbf{65.10} & \textbf{86.62} && 78.50 & 84.14 && 77.62 & 84.43 && 73.96 & 85.55 \\
     \rowcolor[gray]{0.90}
    L-MCM (ours)& 41.80 & 92.46 && \textbf{51.52} & \textbf{89.24} && 85.68 & 74.87 && 75.50 & 77.92  && 72.02 & 84.21 && 65.30 & 83.74 \\
    \rowcolor[gray]{0.90}
    GL-MCM (ours) & \textbf{34.72} & \textbf{94.45} && 61.96 & 88.49 && 72.08 & 84.53 && \textbf{70.66} & \textbf{84.49} && \textbf{71.62} & \textbf{87.72} && \textbf{62.21} & \textbf{87.94}  \\        
    \bottomrule
    \end{tabular}
    }
\vspace{-2mm}
\subcaption{MS-COCO}
\end{minipage}
\\
\vspace{-2mm}
\begin{minipage}[t]{1.0\linewidth}
\centering
     {\tabcolsep = 0.2mm
    \begin{tabular}{@{}lccccccccccccccccc@{}} \toprule
    & \multicolumn{2}{c}{iNaturalist} && \multicolumn{2}{c}{SUN}  && \multicolumn{2}{c}{Texture} && \multicolumn{2}{c}{IN-22K}  && \multicolumn{2}{c}{COCO} && \multicolumn{2}{c}{\textbf{Average}}\\ 
    \cmidrule(lr){2-3} \cmidrule(lr){5-6} \cmidrule(lr){8-9} \cmidrule(lr){11-12} \cmidrule(lr){14-15} \cmidrule(lr){17-18}
   \textbf{Method} & FPR95$\downarrow$ & AUROC$\uparrow$ && FPR95$\downarrow$ & AUROC$\uparrow$ && FPR95$\downarrow$ & AUROC$\uparrow$ && FPR95$\downarrow$ & AUROC$\uparrow$ && FPR95$\downarrow$ & AUROC$\uparrow$ && FPR95$\downarrow$ & AUROC$\uparrow$ \\ \midrule
     MCM & 8.20 & 98.23 && 28.60 & \textbf{94.68} &&  51.70 & 91.45 && 51.40 & 90.94 && 54.50 & 89.02 && 38.88 & 92.86 \\
     \rowcolor[gray]{0.90}
    L-MCM (ours)& 27.70 & 94.97 && 44.20 & 88.87 && 60.00 & 86.23 && 54.10 & 86.27  && 57.00 & 84.06  && 48.72 & 88.08 \\
    \rowcolor[gray]{0.90}
    GL-MCM (ours) &  \textbf{4.20} &  \textbf{98.71} &&  \textbf{23.10} & 94.66  && \textbf{43.00} & \textbf{92.84} && \textbf{41.00} & \textbf{92.38} && \textbf{44.30} & \textbf{90.48} && \textbf{31.12} & \textbf{93.81}\\
    \bottomrule
    \end{tabular}
    }
\vspace{-2mm}
\subcaption{Pascal-VOC}
\end{minipage}
\vspace{-2mm}
\caption{\small{\textbf{Results on MS-COCO and Pascal-VOC.} We find that GL-MCM outperforms MCM in most settings.}}
\label{table:coco_voc}
\end{table*}

\begin{figure}[t]
  \centering
  \includegraphics[keepaspectratio, scale=0.34]
  {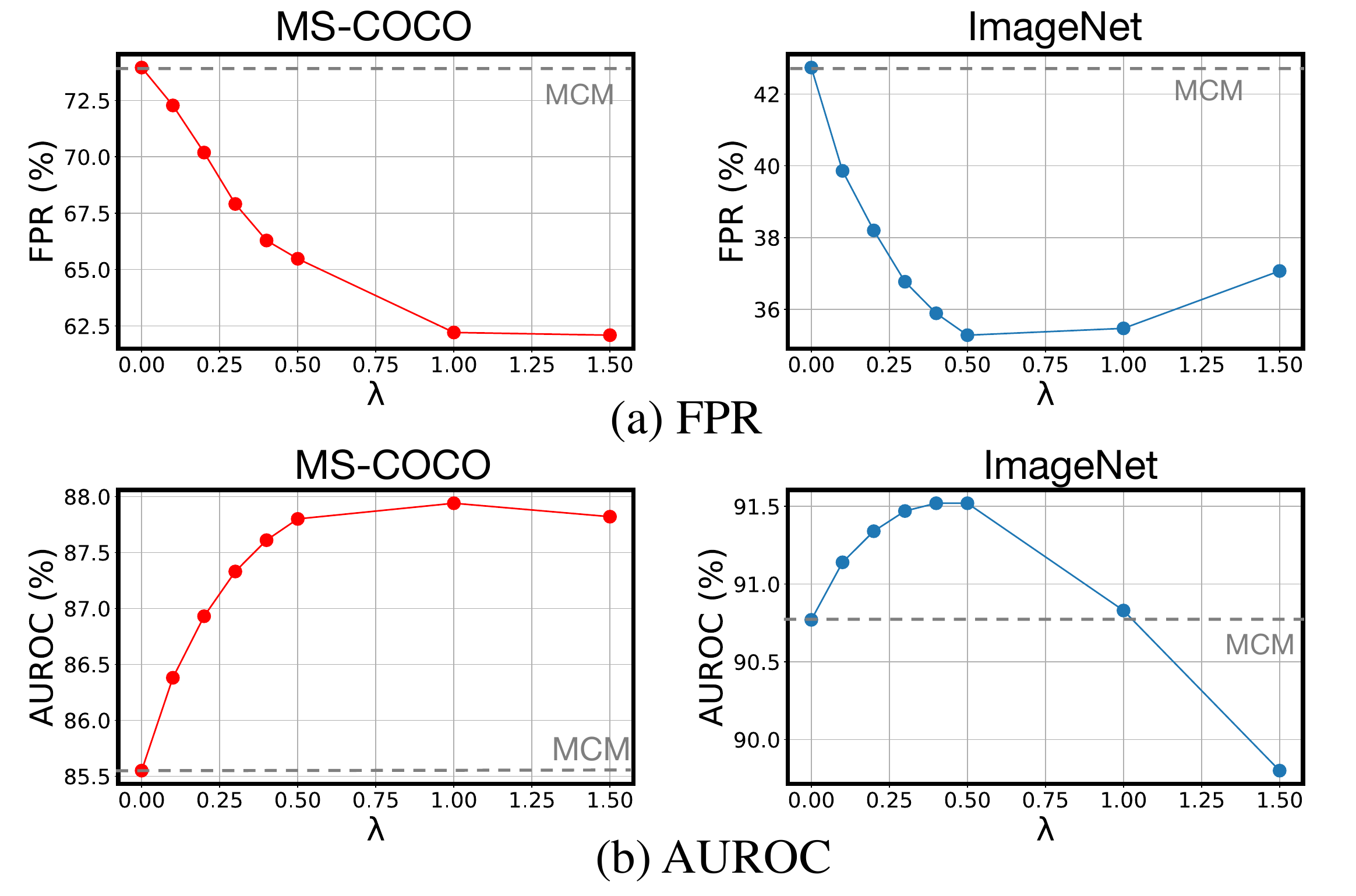}
  \caption{\textbf{Ablation studies on $\lambda$ in Eq.~(\ref{eq:gl_mcm}).} Users can control the type of ID data they aim to detect by changing $\lambda$. A larger $\lambda$ for MS-COCO-like images with both ID and OOD objects and a smaller $\lambda$ for ImageNet-like images with dominant ID objects.}
  \label{fig:ablation_lambda}
\end{figure}

\section{Experiment}
\subsection{Experimental Detail}
\noindent\textbf{ID Datasets.}
We use two types of ID datasets. First is the ImageNet-1K dataset~\citep{deng2009imagenet}, common OOD detection benchmark~\citep{huang2021mos}. The second is MS-COCO~\citep{lin2014microsoft}, Pascal-VOC~\citep{everingham2009pascal}, multi-object datasets. For multi-object datasets, we split annotated classes into ID and OOD classes and created the dataset by collecting images that contain single-class ID objects and one or more class OOD objects in one image. For our datasets, MS-COCO contains 5,000 images with 60 ID classes and 20 OOD classes, and Pascal-VOC contains 1,000 with 14 ID classes and 6 OOD classes. 

\noindent\textbf{OOD Datasets.}
For ImageNet, we use iNaturalist~\citep{van2018inaturalist}, SUN~\citep{xiao2010sun}, PLACES~\citep{zhou2017places}, and TEXTURE~\citep{cimpoi2014describing} as OOD, following the existing work~\citep{huang2021mos}.
For MS-COCO and Pascal-VOC, we use iNaturalist~\citep{van2018inaturalist}, SUN~\citep{xiao2010sun}, TEXTURE~\citep{cimpoi2014describing} and ImageNet-22K~\citep{ridnik2021imagenet} following existing work~\citep{ming2022delving, huang2021mos, wang2021can}. We also use multi-label OOD datasets subset of images from Pascal-VOC (when ID is MS-COCO) and MS-COCO (when ID is Pascal-VOC). 

\noindent\textbf{Setup.} Following existing studies~\citep{tao2023non, ming2022delving}, we use ViT-B/16~\citep{dosovitskiy2020image} as a backbone. Specifically, we use the publicly available CLIP-ViT-B/16 models (\url{https://github.com/openai/CLIP}).  The resolution of CLIP's feature map is 14 $\times$ 14 for CLIP-ViT-B/16. The value of $\tau$ is set to 1.0 from the ablation study in Table~\ref{table:ablation_tau}.

For $\lambda$ in Eq.~(\ref{eq:gl_mcm}), we use 0.5 for ImageNet and 1.0 for COCO and VOC. The rationale behind this choice is that for primarily ID object-centric images like those in ImageNet, it is preferable to prioritize the global score. Conversely, for datasets that are not ID object-centric, such as COCO and VOC, it is more effective to increase the weight of the local score.

\noindent\textbf{Baseline}
We examine the performance of GL-MCM not only in the zero-shot setting but also in the few-shot setting, to confirm the scalability of GL-MCM.
We use MCM with CLIP as a baseline in zero-shot settings. For few-shot settings, we use MCM with CoOp~\citep{zhou2021learning, ming2024does} and LoCoOp~\citep{miyai2023locoop} as baselines.

\noindent\textbf{Comparison Methods.}
To evaluate the effectiveness of our GL-MCM, we compare it with zero-shot detection methods, and fully-supervised detection methods. For zero-shot OOD detection methods, we use MCM~\citep{ming2022delving}. For fully-supervised detection methods, we compare our methods with the following competitive baselines: Fort~\etal./MSP (ViT-B)~\citep{fort2021exploring}, Energy~\citep{liu2020energy}, ODIN~\citep{liang2017enhancing}, ViM~\citep{haoqi2022vim}, KNN~\citep{sun2022out}, and NPOS~\citep{tao2023non}.

In recent years, various zero-shot OOD detection methods beyond MCM have been proposed, including ZeroOE~\citep{fort2021exploring}, ZOC~\citep{esmaeilpour2022zero}, NegLabel~\citep{jiang2024negative}, EOE~\citep{cao2024envisioning}, and CLIPN~\citep{wang2023clipn}. However, these approaches require additional training~\citep{wang2023clipn} and additional negative prompts~\citep{fort2021exploring, esmaeilpour2022zero, jiang2024negative, cao2024envisioning}, which reduces their scalability and flexibility, making them difficult to apply to other methods (e.g., few-shot methods). In this study, we prioritize scalability and flexibility and adopt MCM, the most widely used method as a baseline and comparison method.

\noindent\textbf{Evaluation Metrics.}
For evaluation, we use the following metrics:
(1) the false positive rate of OOD images when the true positive rate of in-distribution images is at 95\% (FPR95), (2) the area under the receiver operating characteristic curve (AUROC).

\begin{figure*}[t]
  \centering
  \includegraphics[keepaspectratio, scale=0.46]
  {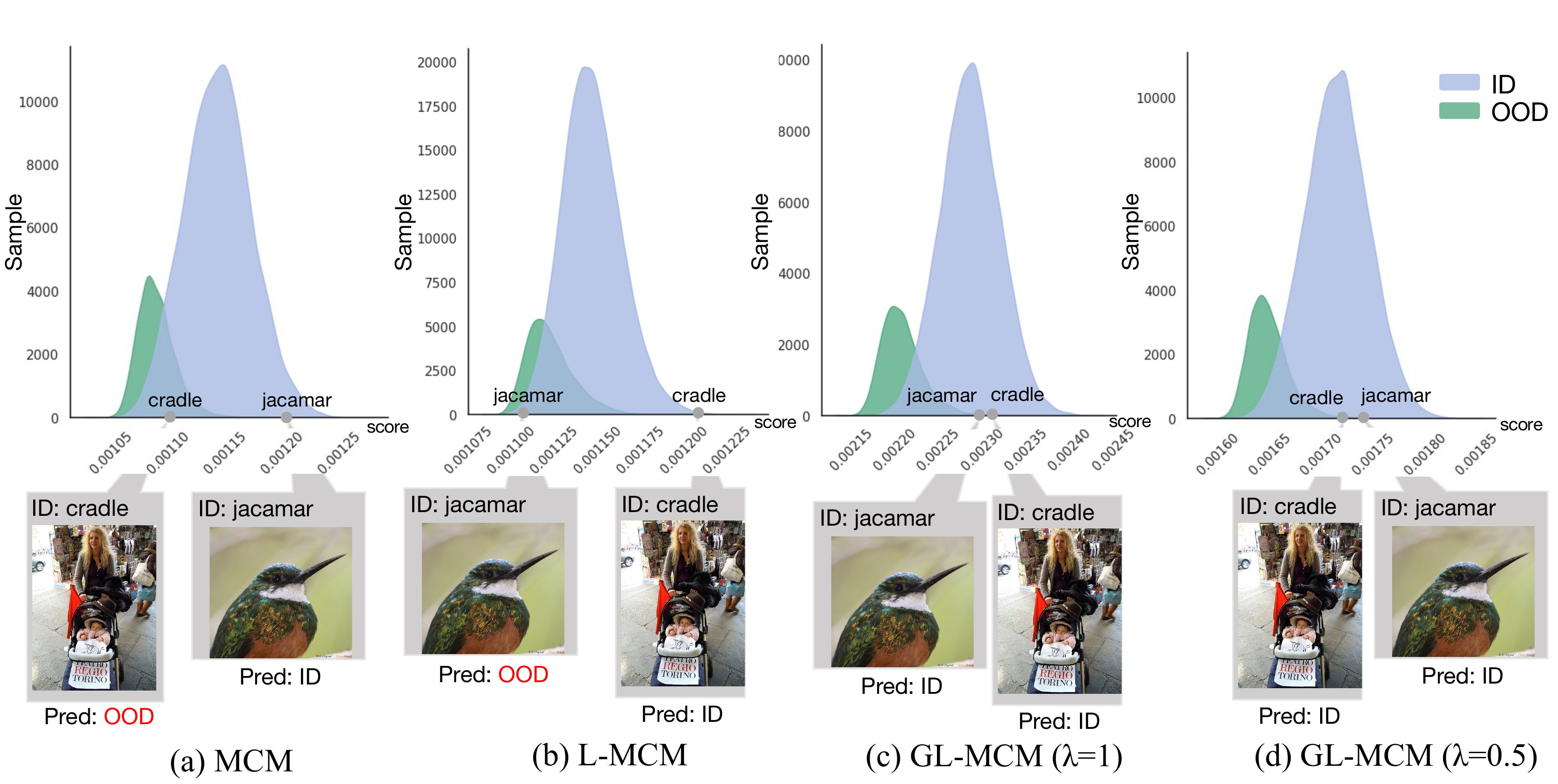}
  \caption{\small{\textbf{Comparison of the histograms of the scores.} We use ImageNet (ID) and iNaturalist (OOD). We use CLIP-ViT-B/16. GL-MCM can identify the ID images, which either MCM or L-MCM can mistake. Furthermore, as shown in (c) and (d), changing $\lambda$ can easily control the score order.}}
  \label{fig:histogram}
\end{figure*}

\subsection{Main Results}
\label{subsec:main_results}
Table~\ref{table:single_IN} summarizes the main results on ImageNet benchmarks.
Table~\ref{table:coco_voc} shows the zero-shot results with the COCO and VOC benchmarks. The findings from these experiments are as follows:

\noindent\textbf{GL-MCM Outperforms MCM on ImageNet.}
In Table~\ref{table:single_IN}, GL-MCM outperforms MCM in most settings (except for Texture). 
This indicates that even in datasets assumed as a single object dataset such as ImageNet, many images unexpectedly include OOD objects, highlighting the importance of considering the presence of these OOD objects. The performance degradation on Textures for GL-MCM is discussed further in Analysis~\ref{subsec:analysis}.

\noindent\textbf{GL-MCM Bootstraps the Performance of Prompt Learning Methods.}
We also found that GL-MCM bootstraps the performance of prompt learning methods such as CoOp and LoCoOp. In particular, we found that GL-MCM is highly compatible
with LoCoOp. This is because LoCoOp integrates a local enhancement loss during training that regularizes local information as OOD. Consequently, we found that GL-MCM is a highly scalable method that can be applied to existing few-shot learning techniques.

\noindent\textbf{GL-MCM is Comparable to Fully Supervised Methods.}
As for the comparison with supervised learning models in Table~\ref{table:single_IN}, the zero-shot performance of GL-MCM with ViT-B is comparable to Energy~\citep{liu2020energy} and superior to other methods. Also, GL-MCM with few-shot learning methods outperforms all comparison methods.
Considering that supervised learning requires fine-tuning with the entire training data, these GL-MCM's results are very satisfactory.

\noindent\textbf{GL-MCM Outperforms MCM on COCO and VOC.}
Table~\ref{table:coco_voc} indicates that GL-MCM also outperforms MCM on COCO and VOC. In addition to the performance improvement, we can confirm that GL-MCM is a good ensemble method to reinforce each other's weaknesses of MCM and L-MCM. As for the average AUROC scores, the superiority of AUROC scores with MCM and L-MCM depends on the datasets, but GL-MCM outperforms them in all settings.

\subsection{Analysis}
\label{subsec:analysis}
\textbf{Flexibility of GL-MCM.}
Fig.~\ref{fig:ablation_lambda} and ~\ref{fig:histogram} demonstrate the high flexibility of GL-MCM. In Fig.~\ref{fig:ablation_lambda}, we show the ablation studies on $\lambda$ in Eq.~(\ref{eq:gl_mcm}). We find that $\lambda$=1 is effective for detecting images with ID and OOD objects (MS-COCO), and $\lambda$=0.50 is effective for detecting images with dominant ID objects (ImageNet). This result indicates the high flexibility of GL-MCM, and users can select the appropriate $\lambda$ according to the application.

In Fig.~\ref{fig:histogram}, we show the histograms of the scores. For the image where the ID object appears locally (\eg cradle), MCM incorrectly judges the image as an OOD image. On the other hand, if the ID object appears largely (\eg jacamar), L-MCM incorrectly judges the ID image as an OOD image. However, since GL-MCM compensates for the shortcomings of both MCM and L-MCM, GL-MCM can correctly identify both images as ID images. In addition, we can see that the order of the scores for cradle and jacamar is reversed for GL-MCM ($\lambda=0.5$) and GL-MCM ($\lambda=1.0$). This shows the high controllability of GL-MCM.

\noindent\textbf{Bootstrapping with Other Localization Methods.}
We investigated the extent of the performance improvement by combining GL-MCM with state-of-the-art localization models. We adopt SAN~\citep{Xu_2023_CVPR}, a state-of-the-art CLIP-based localization method, to get local scores. In SAN, \texttt{[SLS]} tokens have local information, so we replace $\mathbf{x^{\prime}}_i$ in Eq.~(\ref{eq:l_mcm}) with \texttt{[SLS]} tokens. 
As for the comparison method, we add CLIPN~\citep{wang2023clipn_rebuttal}, a state-of-the-art zero-shot OOD detection method. Here, note that CLIPN requires pre-training additional `no' text encoder with a massive dataset (\ie CC3M~\citep{sharma2018conceptual}), so CLIPN is less scalable than GL-MCM. 
From the results in Table~\ref{table:bootstrap_localization}, we find that
GL-MCM outperforms CLIPN for VOC, but it is slightly inferior to CLIPN  for ImageNet. 
Given that CLIPN and GL-MCM are entirely different approaches, the GL-MCM's findings are valid for ImageNet benchmarks. Also, in terms of parameters, using adapters such as SAN (8.4M) is more lightweight than CLIPN with one additional text encoder (63M). Therefore, simply by combining it with an off-the-shelf localization method, GL-MCM can achieve performance close to that of state-of-the-art methods like CLIPN, which require additional large-scale training.
\label{subsec:ablation_study}

\begin{table}[t]
\small
\centering
{\tabcolsep = 0.2mm
\begin{tabular}{@{}lcccccc@{}} \toprule
    & \multicolumn{2}{c}{\textbf{ID: COCO}} & \multicolumn{2}{c}{\textbf{ID:VOC}} & \multicolumn{2}{c}{\textbf{ID:IN-1K}}\\
  \cmidrule(lr){2-3} \cmidrule(lr){4-5} \cmidrule(lr){6-7} 
   \textbf{$\tau$} & \scalebox{0.9}{FPR95}$\downarrow$ & \scalebox{0.9}{AUROC}$\uparrow$ & \scalebox{0.9}{FPR95}$\downarrow$ & \scalebox{0.9}{AUROC}$\uparrow$  & \scalebox{0.9}{FPR95}$\downarrow$ & \scalebox{0.9}{AUROC}$\uparrow$\\ 
   \midrule
   0.01$^\ast$ & 64.89 & 86.70  & 43.92 & 90.80  & 50.72 & 84.70  \\
   \rowcolor[gray]{0.90}
   1.0 & \textbf{62.21}  & \textbf{87.94} & \textbf{31.12} & \textbf{93.81} & \textbf{35.25} & \textbf{91.51} \\
    \bottomrule
    \end{tabular}
    }
\caption{\small{Ablation studies on $\tau$ in Eq.~(\ref{eq:l_mcm}). $\ast$: 0.01 is commonly used for normal classification~\citep{radford2021learning} and segmentation~\citep{Xu_2023_CVPR}.}}
\label{table:ablation_tau}
\end{table}

\begin{table}[t]
\small
\centering
{\tabcolsep = 0.5mm
\begin{tabular}{@{}lcccc@{}} \toprule
    & \multicolumn{2}{c}{\textbf{ID: VOC}}  & \multicolumn{2}{c}{\textbf{ID:IN-1K}}\\
  \cmidrule(lr){2-3} \cmidrule(lr){4-5}
   \textbf{Method} & \scalebox{0.9}{FPR95}$\downarrow$ & \scalebox{0.9}{AUROC}$\uparrow$  & \scalebox{0.9}{FPR95}$\downarrow$ & \scalebox{0.9}{AUROC}$\uparrow$\\ 
 \midrule
   CLIPN  & 21.82 & 95.97 & \textbf{31.10} & \textbf{93.10}  \\
   \hdashline
   SAN ($\tau=0.01$) & 21.48 & 94.64 & 69.81 & 73.37 \\
   \rowcolor[gray]{0.90}
   L-MCM with SAN & \underline{16.40} & \underline{96.31} & 48.29 & 85.96 \\
   \rowcolor[gray]{0.90}
   GL-MCM with SAN & \textbf{16.26} & \textbf{96.42} & \underline{32.91} & \underline{92.10} \\
    \bottomrule
    \end{tabular}
    }
\caption{\small{Experimental Results with an off-the-shell localization method. GL-MCM with SAN~\citep{Xu_2023_CVPR} outperforms CLIPN~\citep{wang2023clipn} a state-of-the-art OOD detection method for VOC and is slightly inferior to CLIPN for ImageNet.}}
\label{table:bootstrap_localization}
\end{table}

\begin{table*}[t]
\centering
\footnotesize
\caption{Comparison results with CLIP-ResNet-50. }
\label{table:comparison_resnet}
\footnotesize
\small
\centering
     {\tabcolsep = 0.8mm
    \begin{tabular}{@{}lcccccccccccccc@{}} \toprule
    & \multicolumn{2}{c}{iNaturalist} && \multicolumn{2}{c}{SUN} && \multicolumn{2}{c}{Places}  && \multicolumn{2}{c}{Texture} && \multicolumn{2}{c}{\textbf{Average}} \\ 
    \cmidrule(lr){2-3} \cmidrule(lr){5-6} \cmidrule(lr){8-9} \cmidrule(lr){11-12} \cmidrule(lr){14-15}
   \textbf{Method} & \scriptsize FPR95$\downarrow$ & \scriptsize AUROC$\uparrow$ && \scriptsize FPR95$\downarrow$ & \scriptsize AUROC$\uparrow$ && \scriptsize FPR95$\downarrow$ & \scriptsize AUROC$\uparrow$ && \scriptsize FPR95$\downarrow$ & \scriptsize AUROC$\uparrow$ && \scriptsize FPR95$\downarrow$ & \scriptsize AUROC$\uparrow$  \\ \midrule
   MCM & \textbf{31.98} & \textbf{93.86} && 46.09 & 90.75 && 60.56 & 85.67 && 60.00 & 85.72 && 49.66 & 89.00 \\
   \rowcolor[gray]{0.90}
   GL-MCM & 32.12 & 93.31 && \textbf{36.40} & \textbf{92.17} && \textbf{48.81} & \textbf{87.51} && \textbf{46.12} & \textbf{88.36} && \textbf{40.86} & \textbf{90.34} \\
    \bottomrule
    \end{tabular}
    }
\end{table*}

\begin{table}[t]
\footnotesize	
\centering
{\tabcolsep = 1.0mm
\begin{tabular}{@{}lcccc@{}} \toprule
   \textbf{Method} & \begin{tabular}{c}Multi \\ input \end{tabular} & \begin{tabular}{c} Time \\ (ms) \end{tabular} $\downarrow$ & \begin{tabular}{c} Mem. \\ (GB) \end{tabular} $\downarrow$ & AUROC $\uparrow$ \\ 
 \midrule
   Grounding DINO & \XSolidBrush   & 87.35 & 4.09 & \underline{87.21} \\
   MCM & \Checkmark & \textbf{19.39} & \textbf{2.10} &  85.55\\
   \rowcolor[gray]{0.90}
   GL-MCM (ours) & \Checkmark & \underline{31.53} & \underline{2.14} & \textbf{87.94}  \\
    \bottomrule
    \end{tabular}
    }
\caption{\small{Comparison with a state-of-the-art object detector, Grounding DINO~\citep{liu2023grounding}. We use COCO as ID. Note that Grounding DINO (as well as GLIP~\citep{li2022grounded}) cannot be evaluated on ImageNet benchmarks due to the input restrictions of the number of ID classes. }}
\label{table:comparison_object_detector}
\end{table}

\noindent\textbf{Experiments with ResNet-50.}
The results in Table~\ref{table:comparison_resnet} showed the OOD detection performance with ResNet-50 on the ImageNet benchmarks. 
An interesting finding beyond mere performance improvements is the different performance trends between ViT and ResNet. For example, in the ViT results shown in Table~\ref{table:single_IN}, GL-MCM performs lower on textures, whereas in Table~\ref{table:comparison_resnet}, GL-MCM shows high performance even on textures. Conversely, for iNaturalist, GL-MCM performs worse than MCM with ViT but outperforms MCM with ResNet.
Comparisons of different backbone architectures have often been explored~\citep{fort2021exploring, hendrycks2019scaling, zhang2023openood}, but no conclusions have yet been drawn because of the different performances on each dataset~\citep{zhang2023openood}. Therefore, a thorough analysis of each backbone is important for future work.

\noindent\textbf{Comparison with Foundation Models for Localization.}
Although other foundation models for localization, such as GLIP~\citep{li2022grounded} and Grounding DINO~\citep{liu2023grounding}, are also candidates, the advantages of using CLIP are (i) inference-time efficiency and (ii) the unlimited number of classes that can be handled. 
As for the inference efficiency, Table \ref{table:comparison_object_detector} shows the comparison results with a state-of-the-art object detector Grounding DINO~\citep{liu2023grounding} on COCO. The results show that GL-MCM outperforms Grounding DINO, especially in terms of both faster inference speed and lower GPU memory consumption. The reason for the high inference efficiency is that GL-MCM utilizes the rich local feature of CLIP and performs pixel-level localization with a single feed-forward pass per image.
As for the number of ID classes, recent localization methods such as Grounding DINO and GLIP have limitations on the number of the ID classes, since they combine all ID classes into a single sentence and input it into the text encoder (\textit{i.e.}, the maximum token number is 256). Therefore, we cannot evaluate Grounding DINO with ImageNet OOD benchmarks, which require long text tokens for input. For the above reasons, we find GL-MCM works efficiently without limiting the number of classes.

\section{Related Work}
\label{sec:related_work}
\textbf{Single-modal OOD Detection.} Various approaches have been developed to address OOD detection in computer vision~\citep{ge2017generative, kirichenko2020normalizing, nalisnick2018deep, neal2018open, oza2019c2ae, hendrycks2016baseline, hendrycks2019scaling, huang2021importance, liang2017enhancing, sun2022out, liu2020energy, lee2018simple, haoqi2022vim, yang2022openood} and natural language processing~\citep{hendrycks2020pretrained, podolskiy2021revisiting, xu-etal-2021-unsupervised}.
However, these OOD detection methods assume the use of backbones trained with single-modal data, so they require task-specific training costs.

\noindent\textbf{CLIP-based Zero-shot OOD Detection.}
To overcome the limitations of single-modal supervised OOD detection, CLIP-based zero-shot OOD detection has been proposed~\citep{fort2021exploring, esmaeilpour2022zero,ming2022delving, jiang2024negative, cao2024envisioning}. There are mainly two approaches for zero-shot OOD detection: using OOD labels (known labels~\citep{fort2021exploring} or pseudo-labels~\citep{esmaeilpour2022zero}) or not using OOD labels~\citep{ming2022delving} to calculate the ID confidence score. 
The earliest work, ZeroOE~\citep{fort2021exploring}, used potential OOD labels in CLIP's textual encoder, but relying on predefined OOD labels limits real-world application. To address this, ZOC~\citep{esmaeilpour2022zero} introduced an OOD label generator based on CLIP's visual encoder to produce pseudo-OOD labels. However, for large datasets with many ID classes, the generator often struggles to produce effective OOD labels, reducing performance.
To overcome the problems with methods using OOD labels, MCM, a method that does not use OOD labels, has been proposed. MCM identifies ID and OOD images by calculating the confidence scores with the softmax score.
Despite its simplicity, MCM greatly outperforms the performance of ZOC, which indicates that the methods not using OOD labels are beneficial. 
However, these methods do not consider the real case when images have ID and OOD objects. We consider this real-world situation and propose a method that can detect ID images even if ID objects appear globally or locally.

\noindent\textbf{Local Features of CLIP.}
Existing work~\citep{zhou2022extract} revealed that CLIP has local visual features aligned with textual concepts. For classification tasks, global features are used, which are created by pooling the feature map with a multi-headed self-attention layer~\citep{vaswani2017attention}. However, these features are known to lose spatial information~\citep{subramanian2022reclip}, so they are not suitable for tasks that require spatial information, such as semantic segmentation. To adopt CLIP to segmentation, \cite{zhou2022extract} proposed the CLIP's local visual features aligned with textual concepts, which can be obtained by projecting the value features of the last attention layer into the textual space. Existing studies used CLIP's local features for semantic segmentation~\citep{zhou2022extract} and multi-label classification~\citep{sun2022dualcoop}. Unlike previous studies, we use local features for OOD detection.

\section{Conclusion}
We introduced Global-Local Maximum Concept Matching (GL-MCM) which combines global and local visual-text alignments to identify ID images in complex, multi-object environments. Our approach, incorporating Local Maximum Concept Matching (L-MCM) for enhanced local feature separability, achieves flexibility and effective OOD detection. Experiments on ImageNet and curated multi-object benchmarks (MS-COCO and Pascal-VOC) show that GL-MCM outperforms existing methods and also provides a flexible use case.

\section*{Statements and Declarations}
\textbf{Funding.}
This research has been funded by JST JPMJCR22U4.
\\[2mm]
\noindent\textbf{Data, Material, Code availability.}
All data, material, and code are available via \url{https://github.com/AtsuMiyai/GL-MCM}.
\\[2mm]
\noindent\textbf{Authors' contribution.}
Atsuyuki Miyai handled all processes from research planning and experiments to writing. Qing Yu and Go Irie provided valuable feedback through discussions and manuscript revisions. Kiyoharu Aizawa is the supervisor of Atsuyuki Miyai and also made significant contributions through discussions and manuscript editing.

\bibliography{sn-bibliography}

\maketitle

\newcommand\beginsupplement{%
        \setcounter{table}{0}
        \renewcommand{\thetable}{\Alph{table}}%
        \setcounter{figure}{0}
        \renewcommand{\thefigure}{\Alph{figure}}%
     }
\beginsupplement
\appendix

\section{Additional Experimental Results}\label{secB1}
\label{sec:additional_exp}
\subsection{Ablation of Score Functions.}
To validate the effectiveness of GL-MCM, we experiment with different score functions. We compute score functions in seven ways:
\begin{itemize}
      \item (i) Entropy: the negative entropy of softmax scaled cosine similarities for both global and local scores. Formally, $S_{\operatorname{entropy}} = -(H(\mathbf{p}) + \max_{i}(H(\mathbf{p_i})))$, where 
      $\mathbf{p}$ and $\mathbf{p_i}$ denote the $K$ dimension output probability of the global feature $\mathbf{x}^\prime$ and the local feature $\mathbf{x}^{\prime}_i$ respectively.
      $H(\cdot)$ denotes the entropy function.
      \item (ii) Var: the variance of the cosine similarities for both global and local scores. Formally, $S_{\operatorname{var}} = V(\mathbf{s}) + \max_{i}(V(\mathbf{s}_i))$, where $\mathbf{s}$ and $\mathbf{s}_i$ denote the $K$ dimension cosine similarity of the global feature and the local feature. $V(\cdot)$ denotes the variance function.
      \item (iii) Cos: the maximum cosine similarities for both global and local scores. Formally, $S_{\operatorname{cos}} = \max _{t}(\operatorname{sim}\left(\mathbf{x^{\prime}}, \mathbf{y}_t\right)) + \max _{t, i}(\operatorname{sim}\left(\mathbf{x^{\prime}}_i, \mathbf{y}_t\right))$.
      \item (iv) Global or Local MCM: the higher score of either the global or local confidence score. Formally, $S_{\mathrm{G-or-L-MCM}} = \max(S_{\mathrm{MCM}}, S_{\mathrm{L-MCM}})$
      \item (v) L-Class-Avg CM: the maximum class-wise average concept matching score for local information. Formally, we define the prediction label $\mathrm{pred}_i$ for each region $i$ of the feature map as follows:
      $\mathrm{pred}_i = \argmax_{t} 
      \frac{e^{\operatorname{sim}\left({\mathbf{x^{\prime}}_i}, \mathbf{y}_t\right) / \tau}}{\sum_{c \in \mathcal{T}_{in}} 
      e^{\operatorname{sim}\left({\mathbf{x^{\prime}}}_i, \mathbf{y}_c\right) / \tau}}$. 
      We define the set of the region $i$ where $\mathrm{pred}_i = t\in \mathcal{T}_{in}$ as $I(t)$. With $I(t)$, the score function $S_{\mathrm{L-Class-avg}}$ is defined as follow:
\begin{equation}
\footnotesize
S_{\mathrm{L-Class-Avg}} = \max _{t} \frac{1}{|I(t)|}\cdot \sum_{i\in I(t)} \frac{e^{\operatorname{sim}\left({\mathbf{x^{\prime}}_i}, \mathbf{y}_t\right) / \tau}}{\sum_{c \in \mathcal{T}_{in}} e^{\operatorname{sim}\left({\mathbf{x^{\prime}}_i}, \mathbf{y}_c\right) / \tau}}.
\label{eq:class_avg}
\end{equation}
      \item (vi) L-Class-Avg CM + MCM: the ensemble method with L-Class-AVG CM and MCM. Formally, the score function is $S_{\mathrm{L-Class-Avg}} + S_{\mathrm{MCM}}$.
      \item (vii) L-Max CM + MCM: GL-MCM (ours). Formally, $S_{\mathrm{GL-MCM}} = S_{\mathrm{L-MCM}} + S_{\mathrm{MCM}}$.
\end{itemize}
In Table~\ref{comparison_other_scaling}, we show the results with these score functions. We find that (vi) L-Class-Avg CM + MCM and (vii) GL-MCM give high results compared to other score functions. Since (vi) also uses both global and local scores, it is a method with the same concept as GL-MCM. While (vi) and (vii) use similar scores, (vii) is chosen as the main method due to its ease of execution and implementation.

\begin{table*}[t]
\scriptsize
\centering
{\tabcolsep = 1.5mm
\begin{tabular}{@{}lcccccc@{}} \toprule
     & \multicolumn{2}{c}{\textbf{ID:ImageNet}} & \multicolumn{2}{c}{\textbf{ID: COCO}} & \multicolumn{2}{c}{\textbf{ID:VOC}} \\
  \cmidrule(lr){2-3} \cmidrule(lr){4-5} \cmidrule(lr){6-7} 
   \textbf{Method} & FPR95$\downarrow$ & AUROC$\uparrow$ & FPR95$\downarrow$ & AUROC$\uparrow$  & AUROC$\uparrow$ & FPR95$\downarrow$ \\ 
 \midrule
    (i) Entropy  & 79.95 & 68.76 & 80.03 & 76.23 & 55.16 & 90.17 \\
    (ii) Var & 80.50 & 68.08 & 80.56 & 75.38 & 44.46 & 91.35  \\
    (iii) Cos & 53.1 & 86.11 & 90.86 & 80.11 & 55.16 & 90.17  \\
    (iv) Global or Local MCM & 51.5 & 85.29 & \underline{63.38} & 84.96 & 47.30 & 89.15 \\
    (v) L-Class-Avg CM & 74.72 & 73.49 & 64.27 & 81.58 & 59.70 & 86.04 \\
    (vi) L-Class-Avg CM + MCM & \textbf{33.62} & \textbf{91.84}  & 63.62 & \textbf{87.97} & \underline{37.94} & \underline{93.36} \\
    (vii) L-MCM + MCM (GL-MCM) & \underline{35.25} & \underline{91.51} & \textbf{62.21} & \underline{87.94} & \textbf{31.12} & \textbf{93.81}  \\ 
    \bottomrule
    \end{tabular}
    }
\caption{Comparison with other score functions. We use CLIP-B/16 as the backbone.}
\label{comparison_other_scaling}
\end{table*}

\subsection{Analysis of Extracted ID Data}
We analyze the extracted ID data with MCM and GL-MCM on MS-COCO. 
To extract ID data, we set a threshold to 0.0178 for the MCM score and 0.0358 for the GL-MCM score in this experiment (the number of OOD data within these thresholds is about the same.). Table~\ref{table_num_ID_coco} indicates categories where there is a difference of 30 or more in the number of ID data extracted between GL-MCM and MCM. These categories are often pictured with other objects, so they cannot be detected as ID images by MCM, but they can be detected as ID images by our GL-MCM. Besides, we find that the number of ID images with GL-MCM is equal or larger than the number with MCM in all categories. 

\begin{figure*}[t]
  \centering
  \includegraphics[keepaspectratio, scale=0.5]
  {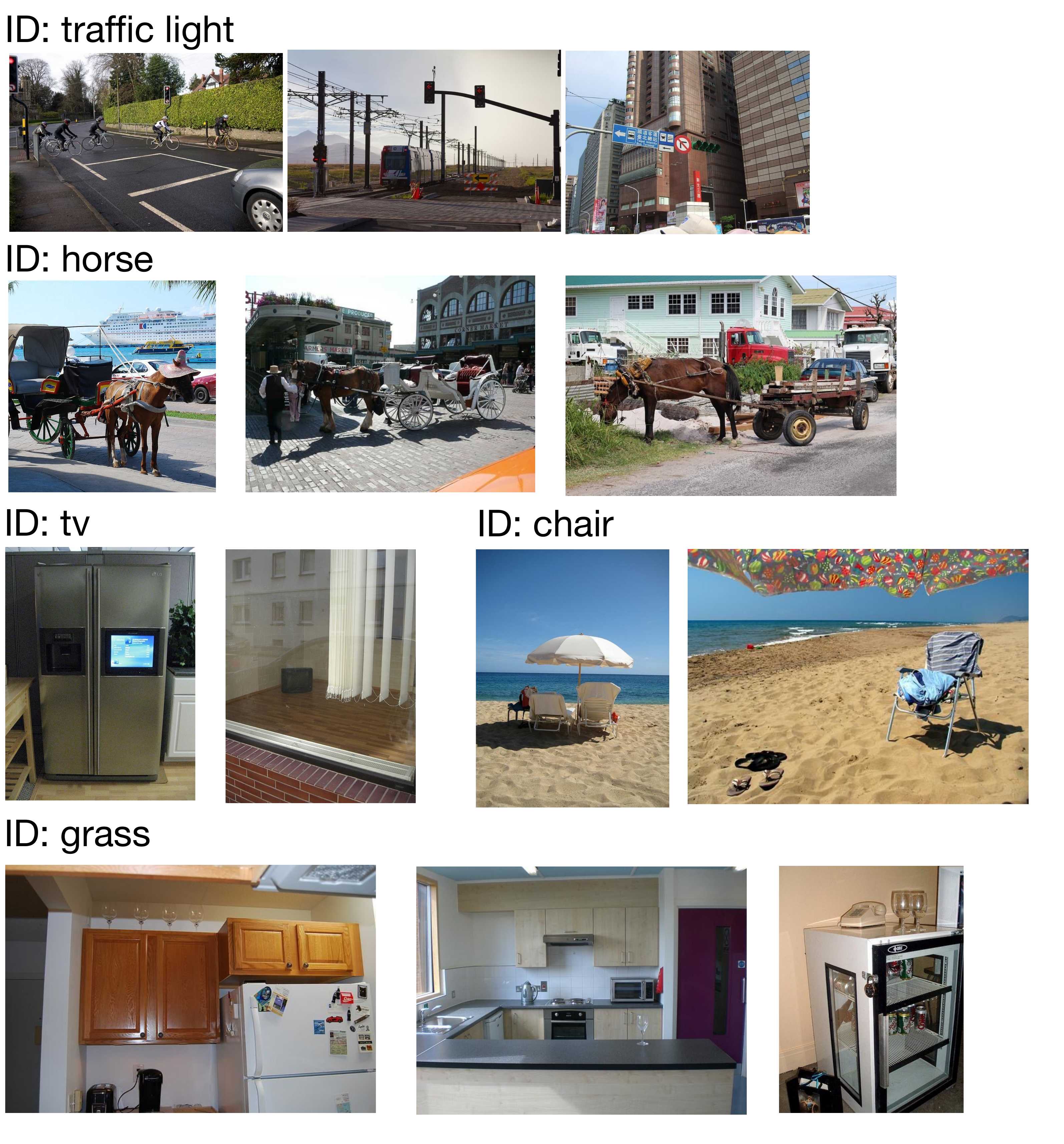}
  \caption{Samples of extracted ID data on MS-COCO. These samples were not extracted by MCM, but were correctly extracted by GL-MCM.} 
  \label{fig:ID_sample}
\end{figure*}

In Fig.~\ref{fig:ID_sample}, we show examples of extracted ID data by GL-MCM. These samples were not extracted by MCM because extra OOD objects appear in images. 

\subsection{Visualization of Alignment Maps}
\begin{figure}[t]
  \centering
  \includegraphics[keepaspectratio, scale=0.42]
  {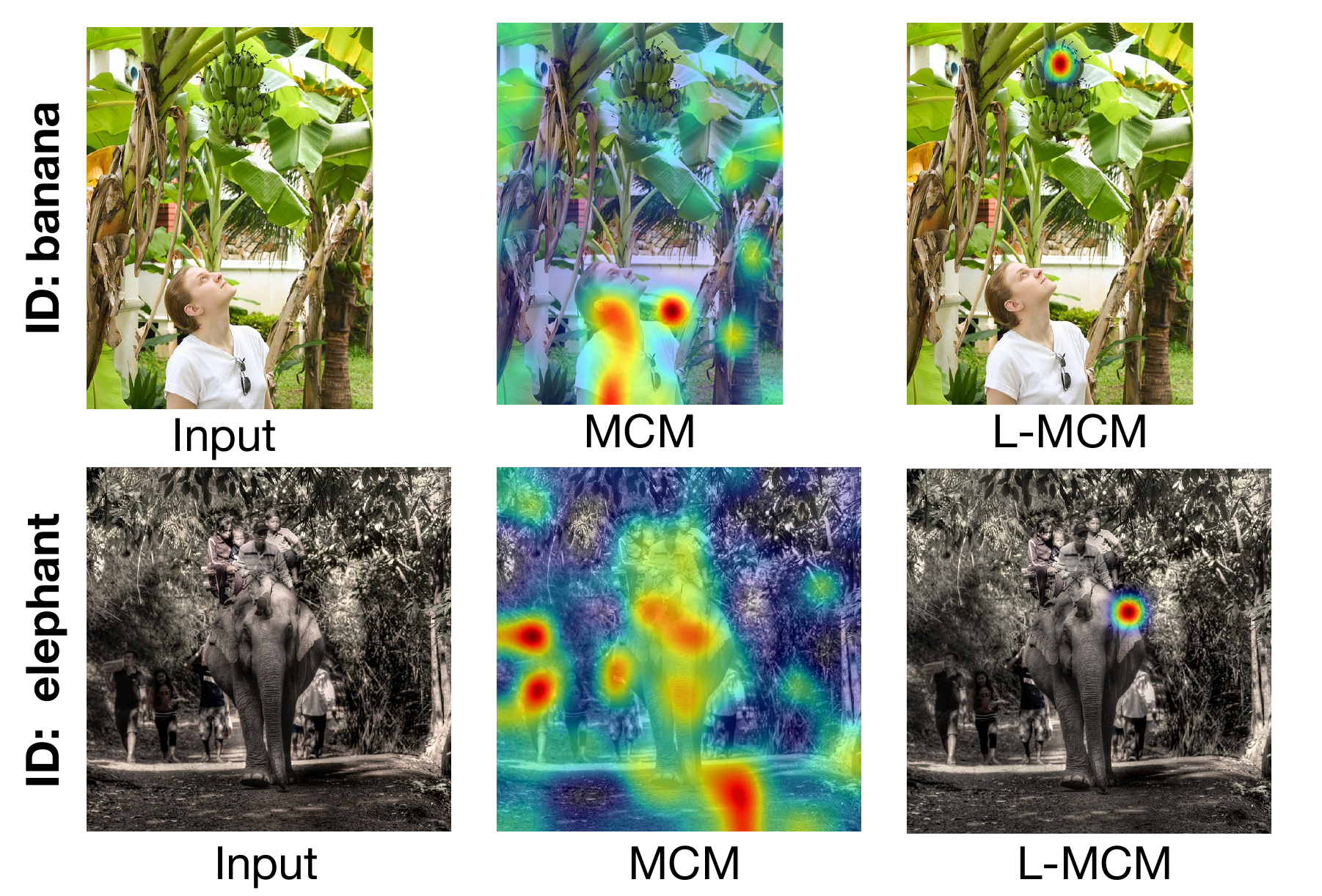}
  \caption{\textbf{Visualization of the alignment maps} of MCM and L-MCM. }
  \label{fig:visualize}
\end{figure}

In Fig.~\ref{fig:visualize}, we visualize the alignment maps for MCM and L-MCM. Specifically, for MCM, we show the alignment map of the CLIP-B/16 encoder, and for L-MCM, we show the region with the local maximum concept matching score. Note that it is not possible to visualize the attention map of GL-MCM, but it is regarded as a combination of both these maps. In the case of an ID object captured locally, MCM does not pool its features well, as shown at the top of Fig.~\ref{fig:visualize}. On the other hand, L-MCM cannot adequately take into account the features of objects captured globally, as shown at the bottom of Fig.~\ref{fig:visualize}. Therefore, GL-MCM, which incorporates these two approaches, can reinforce these two weaknesses.

\begin{table}[t]
\centering
\footnotesize
\centering
\centering
    \begin{tabular}{@{}lccc@{}} \toprule

    & \multicolumn{2}{c}{\#ID}  & \\  
    \cmidrule(lr){2-4}
    Category & MCM & GL-MCM & diff. \\ 
     \midrule 
     all & 3,247 & 3,817 & 570 \\
     \midrule 
     traffic light & 279 & 412 & 133  \\
     cat & 279 & 347 & 68  \\ 
     dog & 152  & 197 & 45 \\     
     bottle & 77 & 109 & 32  \\ 
    \bottomrule
    \end{tabular}
\caption{\textbf{The number of extracted ID data on MS-COCO.} We reported numbers for categories with a difference of 30 or more between GL-MCM and MCM.}
\label{table_num_ID_coco}
\end{table}

\begin{table*}[t]
\centering
\footnotesize
\centering
\centering
     {\tabcolsep = 0.5mm
    \begin{tabular}{@{}lccccccccccccccccc@{}} \toprule
    & \multicolumn{2}{c}{iNaturalist} && \multicolumn{2}{c}{SUN}  && \multicolumn{2}{c}{Texture} && \multicolumn{2}{c}{IN-22K} && \multicolumn{2}{c}{VOC} && \multicolumn{2}{c}{Average}\\ 
    \cmidrule(lr){2-3} \cmidrule(lr){5-6} \cmidrule(lr){8-9} \cmidrule(lr){11-12} \cmidrule(lr){14-15} \cmidrule(lr){17-18}
   \textbf{Method} & FPR95$\downarrow$ & AUROC$\uparrow$ && FPR95$\downarrow$ & AUROC$\uparrow$ && FPR95$\downarrow$ & AUROC$\uparrow$ && FPR95$\downarrow$ & AUROC$\uparrow$ && FPR95$\downarrow$ & AUROC$\uparrow$ && FPR95$\downarrow$ & AUROC$\uparrow$ \\ \midrule
   \multicolumn{11}{l}{\textbf{ResNet-50}}\\
    MCM & 38.78 & 92.81 && 36.16 & 89.85 && 47.68 & 88.39 && 69.36 & 80.58 && 49.05 & 88.53 && 48.21 & 88.03 \\
    L-MCM (ours)& 38.14 & 94.23 && 44.6 & 90.62  && 35.74 & 90.73 && 63.80 & 86.75 && 55.30 & 88.73 && 47.53 & 90.21 \\
    GL-MCM (ours) & \textbf{20.54} & \textbf{96.27} && \textbf{27.44} & \textbf{92.75} && \textbf{29.10} & \textbf{92.19} && \textbf{55.76} & \textbf{87.76} && \textbf{39.20} & \textbf{91.78} && \textbf{34.41} & \textbf{92.15} \\
     \midrule
    \multicolumn{11}{l}{\textbf{ViT-B}}\\
     MCM & 45.84 & 92.2 && 74.94 & 81.19 &&  \textbf{54.54} & \textbf{86.51} && 68.24 & 84.36 && 67.92 & 83.83 && 62.3 & 85.62 \\
    L-MCM (ours)& 23.12 & 95.08 && \textbf{32.88} & \textbf{92.17} && 70.24 & 77.85 && 58.46 & 81.25  && 51.85 & 87.6 && 47.31 & 86.79 \\
    GL-MCM (ours) & \textbf{18.90} & \textbf{96.46} && 43.94 & 90.59 && 55.98 & 85.92 && \textbf{54.96} & \textbf{86.31} && \textbf{51.30} & \textbf{89.78} && \textbf{45.02} & \textbf{89.81} \\        
    \bottomrule
    \end{tabular}
    }
\caption{\textbf{Results for the ID samples with multi-class ID objects.} We use the ID samples containing multi-class ID objects and OOD objects from MS-COCO. }
\label{table_multi_ID_coco}
\end{table*}

\subsection{Results on ID Images with Multi-class ID and OOD Objects}
\label{subsec:multi_class_id_objects}
In this section, we conduct experiments on the ID dataset where each ID image has multi-class ID objects and some OOD objects. In the main paper, we used ID datasets where each ID image has single-class ID objects and some OOD objects and did not examine the case where each ID image has multi-class ID objects and OOD objects. We create the ID dataset containing multi-class ID objects and some OOD objects from MS-COCO. In Table~\ref{table_multi_ID_coco}, we show the results. From these results, we find that GL-MCM also outperforms other comparison methods in this setting.

\section{Limitations}\label{secC1}
\label{sec:limitaion}
\textbf{ID Classification.}
ID detection is a task of distinguishing ID and OOD, which is the same as OOD detection and does not focus on improving close-set ID recognition accuracies.
However, this task can reduce considerable human effort and help annotators even without the ability to close-set classifications.

\vspace{2mm}
\noindent\textbf{Need of Local Visual-text Alignment.}
This method relies on models with strong local visual-text alignment capabilities, such as CLIP's Image Encoder. Therefore, it is difficult to apply this method to models that do not have such local visual-text alignment. However, since many methods are currently being proposed based on CLIP, our work provides a beneficial insight into the research of vision-language research fields.

\section{Dataset Details}
\label{dataset_detail}
\subsection{ID Dataset}
We created ID datasets from ImageNet, MS-COCO, and Pascal-VOC. The detailed information on these datasets is as follows:
\\[1mm]
\textbf{ImageNet.}
We use the ImageNet validation dataset as ID data. ImageNet validation dataset has 50,000 images with 1,000 classes. We use this dataset to validate the effectiveness of our proposal on common single-object OOD benchmarks.
\\[1mm]
\textbf{MS-COCO.}
Our MS-COCO dataset contains 5,000 images with 60 ID classes and 20 OOD classes from MS-COCO2017 training images. As for the split between ID and OOD, we mainly split them so as not to have any overlaps of supercategories between ID and OOD samples. In addition, the classes ``person" and ``potted plant" are also used as OOD because there are many samples containing such objects in the OOD datasets. Specifically, for ID classes, we use ``traffic light", ``fire hydrant", ``stop sign", ``parking meter", ``bench", ``bird", ``cat", ``dog", ``horse", ``sheep", ``cow", ``elephant", ``bear", ``zebra", ``giraffe", ``frisbee", ``skis", ``snowboard", ``sports ball", ``kite", ``baseball bat", ``baseball glove", ``skateboard", ``surfboard", ``tennis racket", ``bottle", ``wine glass", ``cup", ``fork", ``knife", ``spoon", ``bowl", ``banana", ``apple", ``sandwich", ``orange", ``broccoli", ``carrot", ``hot dog", ``pizza", ``donut", ``cake", ``chair", ``couch", ``bed", ``dining table", ``toilet", ``tv", ``laptop", ``mouse", ``remote", ``keyboard", ``cell phone", ``book", ``clock", ``vase", ``scissors", ``teddy bear", ``hair drier", ``toothbrush". 
For OOD classes, we use ``person", ``bicycle", ``car", ``motorcycle", ``airplane", ``bus", ``train", ``truck", ``boat", ``backpack", ``umbrella", ``handbag", ``tie", ``suitcase", ``potted plant", ``microwave", ``oven", ``toaster", ``sink", ``refrigerator".
In this supplementary experiment, we use the dataset where each ID image has multi-class ID objects and OOD objects. For multi-class ID objects, we exclude ``baseball bat" and ``scissors" from the above ID classes due to the creating process.
\\[1mm]
\textbf{Pascal-VOC.}
Our Pascal-VOC dataset contains 1,000 with 14 ID classes and 6 OOD classes. For ID classes, we use ``aeroplane", ``bicycle", ``bird", ``boat", ``bottle", ``cow", ``diningtable", ``dog", ``horse", ``motorbike",  ``sheep", ``sofa", ``train", ``tvmonitor``. For OOD classes, we use ``pottedplant", ``chair", ``cat", ``car", ``bus", ``person"

\subsection{OOD Dataset}
\textbf{iNaturalist.}
We use iNaturalist as OOD data for all ID datasets. We use the iNaturalist dataset provided by MOS~\citep{huang2021mos}. We exclude the plant-related class from ID datasets, so the class overlap does not happen.
\\[1mm]
\textbf{SUN.}
We use SUN as OOD data for all ID datasets. We use the SUN dataset provided by MOS~\citep{huang2021mos}. We exclude the plant-related class from ID datasets, so the class overlap does not happen.
\\[1mm]
\textbf{Textures.}
We use Textures as OOD data. Texture dataset consists of 5,640 images of textural
patterns. This dataset does not have the overlap of the classes in MS-COCO and Pascal-VOC, and ImageNet.
\\[1mm]
\textbf{ImageNet-22K.}
We use ImageNet-22K as OOD data for MS-COCO and Pascal-VOC following~\citep{wang2021can}.  This dataset is provided without the overlap of the classes in MS-COCO and Pascal-VOC~\citep{wang2021can}.
\\[1mm]
\textbf{Places.}
We use the Places dataset as OOD data when the ID dataset is ImageNet-1K. This dataset is provided without the overlap of the classes in ImageNet-1K~\citep{huang2021mos}.
\\[1mm]
\textbf{Pascal-VOC.}
We use a subset of Pascal-VOC as OOD when the ID dataset is MS-COCO. We use the samples containing the classes of ``aeroplane", ``bicycle", ``boat", ``bus", ``car", ``motorbike", ``person", ``pottedplant", and ``train" classes for this subset not to overlap the ID classes. The subset contains 4,000 images.
\\[1mm]
\textbf{MS-COCO.}
We use a subset of MS-COCO as OOD when the ID dataset is Pascal-VOC. We use the samples not containing the classes of ``bicycle", ``motorcycle", ``airplane", ``train", ``boat", ``bird", ``dog", ``horse", ``sheep", ``cow", ``bottle", ``couch", ``dining table", ``tv" for this subset not to overlap the ID classes. The subset contains 1,000 images.

\end{document}